%% file: main.tex
\documentclass{style/naturep}

\usepackage{amssymb}
\usepackage{amsmath}
\usepackage{graphicx}
\usepackage{algorithmicx}
\usepackage{algorithm}
\usepackage{adjustbox}
\usepackage{graphicx}
\usepackage{bm}

\usepackage{xurl}
\usepackage{enumitem}
\setlist[itemize]{noitemsep, nolistsep}

\usepackage[noend]{algpseudocode}
\usepackage{bbm}
\usepackage{lineno}
\usepackage[margin=0.6in]{geometry}
\usepackage{ragged2e}
\usepackage{setspace}
\usepackage{longtable}
\usepackage{hyperref}
\usepackage{anyfontsize}
\usepackage{setspace}
\usepackage{float}
\usepackage{booktabs}
\usepackage{multirow}
\usepackage{blindtext}
\usepackage{tabularx}
\usepackage{caption}
\usepackage{subcaption}
\usepackage{enumitem, kantlipsum}
\usepackage{xspace}
\usepackage{pifont}
\newcommand{\xmark}{\ding{55}}%

\usepackage{graphicx}
\usepackage{amsmath}
\usepackage{amssymb}
\usepackage{booktabs}
\usepackage{float}
\usepackage{placeins}
\usepackage[table,x11names]{xcolor}
\definecolor{maroon}{cmyk}{0,0.87,0.68,0.32}
\definecolor{gray}{rgb}{0.3,0.3,0.3}

\usepackage{caption}
\usepackage{afterpage}

\usepackage{colortbl} 
\usepackage{booktabs} 
\usepackage{xcolor} 
\definecolor{LightGray}{gray}{0.9}

\newcommand{\ours}{\textsc{MUPAD}\xspace}

\newcommand\Heading[1]{
  \noindent\textbf{\Large{#1}}
}

\newcommand\heading[1]{
  \noindent\textbf{\large{#1}}
}

\title{\begin{flushleft}{\begin{spacing}{1}
   A Generative Foundation Model for Multimodal Histopathology
\end{spacing}}\end{flushleft}}

\makeatletter
\let\saved@includegraphics\includegraphics
\AtBeginDocument{\let\includegraphics\saved@includegraphics}

\makeatother

\begin{document}

\input{sections/0-cover_page}
\clearpage

\Heading{Abstract}
\input{sections/0-abstract}

\begin{spacing}{1.35}
\clearpage
\Heading{Introduction}
\input{sections/1-introduction}


\Heading{Results}
\input{sections/2-results}

\clearpage
\Heading{Discussion}

\input{sections/3-discussion}

\clearpage
\Heading{Online Methods}

\input{sections/4-methods}

\Heading{Data Availability}

\noindent
All datasets used in this study are publicly available or accessible via controlled-access repositories.
For pretraining, H\&E whole-slide images and RNA-seq profiles were sourced from The Cancer Genome Atlas (TCGA; \url{https://portal.gdc.cancer.gov})\cite{Weinsteinetal2013}, the Genotype-Tissue Expression (GTEx) project (\url{https://gtexportal.org})\cite{Lonsdaleetal2013}, PAIP (\url{https://paip2023.grand-challenge.org})\cite{kim2023paip}, and the Prostate, Lung, Colorectal, and Ovarian (PLCO) Cancer Screening Trial (\url{https://cdas.cancer.gov/plco})\cite{Zhuetal2013}.
Text--image pretraining pairs were derived from PathGen-1.6M\cite{sun2024pathgen}.
For evaluation, the following publicly available datasets were used:
the HISTAI dataset\cite{nechaev2025histai} for image-to-image generation;
PathMMU\cite{sun2024pathmmu} for vision--language retrieval;
SkinCancer\cite{skincaner}, PanNuke\cite{gamper2019pannuke}, UniToPatho\cite{barbano2021unitopatho}, LC25000\cite{borkowski2019lung}, and SICAPv2\cite{silva2020going} for few-shot classification;
the MOSAIC spatial transcriptomics dataset (EGA accession EGAC50000000398; \url{https://ega-archive.org/dacs/EGAC50000000398}) for transcriptomics-to-H\&E generation;
TCGA lung and brain cohorts\cite{Weinsteinetal2013} for fresh-frozen to FFPE translation;
HER2Match\cite{HER2match_data,klockner2025gans} and MIST\cite{li2023adaptive} for H\&E-to-IHC translation;
and the ORION colorectal cancer dataset\cite{lin2023high} for H\&E-to-multiplex immunofluorescence translation.
Access to controlled-access datasets (e.g., MOSAIC via EGA, PLCO via CDAS) requires approval from the respective data access committees.

\Heading{Competing interests}

\noindent
The authors declare no competing interests.

\Heading{Code Availability}

\noindent The source code is provided as supplementary software for review and will be made publicly available upon publication.

\end{spacing}

\clearpage

\begin{nolinenumbers}
\Heading{References}

\vspace{2mm}

\begin{spacing}{0.9}
\bibliographystyle{naturemag}
\bibliography{main}
\end{spacing}
\end{nolinenumbers}
\clearpage

\setcounter{figure}{0}
\renewcommand{\figurename}{Extended Data Figure}

\input{appendix/figs}
\clearpage

\setcounter{table}{0}
\renewcommand{\tablename}{Extended Data Table}

\input{appendix/tables}

\clearpage

\end{document}

%% file: sections/0-cover_page.tex
\maketitle
\vspace{-20mm}
\begin{spacing}{1.4}
\noindent 
Jinxi Xiang$^{1,\ddag}$, 
Mingjie Li$^{2,\ddag}$, 
Siyu Hou$^{3}$, 
Yijiang Chen$^{1}$,
Xiangde Luo$^{1}$,
Yuanfeng Ji$^{1}$,
Xiang Zhou$^{3}$,
Ehsan Adeli$^{2,4,6}$,
Akshay Chaudhari$^{6,7,8}$,
Curtis P.\ Langlotz$^{6,7,8}$,
Kilian M. Pohl$^{2,5}$,
Ruijiang Li$^{1*}$
\end{spacing}

\vspace{-7mm}
\begin{spacing}{1.4}
\begin{affiliations}
    \item Department of Radiation Oncology, Stanford University School of Medicine, Stanford, CA
    \item Department of Psychiatry and Behavioral Sciences, Stanford University School of Medicine, Stanford, CA
    \item Department of Statistics and Data Science, Yale University,  New Haven, CT
    \item Department of Computer Science, Stanford University, CA
    \item Department of Electrical Engineering, Stanford University, CA
    \item Department of Biomedical Data Science, Stanford University, Palo Alto, CA
    \item Department of Radiology, Stanford University, Palo Alto, CA
    \item Center for Artificial Intelligence in Medicine and Imaging, Stanford University, Palo Alto, CA
    \\ $\boldsymbol{\ddag}$ Equal contribution
    \\ \textbf{*} Correspondence to: Ruijiang Li (rli2@stanford.edu)
\end{affiliations}
\end{spacing}

\noindent \textbf{Acknowledgments}

\noindent
This work was supported by the National Cancer Institute (R01CA269599, R01CA285456, R01CA290715). K.M. Pohl was supported by the National Institute of Mental Health (R01MH145082), the National Institute of Drug Abuse (R01DA057567), and the Stanford HAI Hoffman-Yee Grant.

\vspace{10mm}

%% file: sections/0-abstract.tex
\begin{spacing}{1.38}
\noindent
Accurate diagnosis and treatment of complex diseases require integrating histological, molecular, and clinical data, yet in practice these modalities are often incomplete owing to tissue scarcity, assay cost, and workflow constraints. Existing computational approaches attempt to impute missing modalities from available data but rely on task-specific models trained on narrow, single source–target pairs, limiting their generalizability. Here we introduce \textbf{\ours} (\textbf{Mu}ltimodal \textbf{P}athology \textbf{D}iffusion), a generative foundation model that embeds hematoxylin and eosin (H\&E)-stained histology, molecular RNA profiles, and clinical text into a shared latent space through a diffusion transformer with decoupled cross-modal attention. Pretrained on 100 million histology image patches, 1.6 million text–histology pairs, and 10.8 million RNA–histology pairs spanning 34 human organs, \ours supports diverse cross-modal synthesis tasks with minimal or no task-specific fine-tuning. For text-conditioned and image-to-image generation, \ours synthesizes histologically faithful tissue architectures, reducing Fréchet inception distance (FID) scores by 50\% relative to domain-specific models and improving few-shot classification accuracy by up to 47\% through synthetic data augmentation. For RNA-conditioned histology generation, \ours reduces FID by 23\% compared with the next-best method while preserving cell-type distributions across five cancer types. As a virtual stainer, \ours translates H\&E images to immunohistochemistry and multiplex immunofluorescence, improving average marker correlation by 37\% over existing approaches. These results demonstrate that a single, unified generative model pretrained across heterogeneous pathology modalities can substantially outperform specialized alternatives, providing a scalable computational framework for multimodal histopathology.
\end{spacing}

\noindent \textbf{Project Page}: \url{https://mupad-demo.vercel.app/}

\noindent \textbf{Code}: The source code is provided as supplementary software for review and will be made publicly available upon publication. 

\noindent \textbf{Model}: \url{https://huggingface.co/collections/xiangjx/mupad-multimodal-pathology-diffusion-model}

%% file: sections/1-introduction.tex
\noindent
Modern pathology diagnosis integrates multiple data modalities~\cite{lipkova2022artificial, moor2023foundation, chen2022pan, xiang2025vision, ding2025multimodal}. For complex diseases such as cancer, clinical decision-making typically requires, at a minimum, standard haematoxylin and eosin (H\&E)-stained histology, supplemented with immunohistochemistry (IHC) or multiplex immunofluorescence (mIF) for protein marker quantification and molecular profiling for genomic or transcriptomic characterization~\cite{swanson2023patterns, chen2022pan, liao2025deep, xu2025multimodal}. In practice, however, acquiring this full complement of data is frequently infeasible: tissue quantities may be insufficient for sequential assays, specialized staining protocols are costly and not universally accessible, and turnaround times can delay clinical decisions~\cite{li2026ai, liu2025spatial}. The resulting gaps in multimodal information contribute to diagnostic uncertainty and can compromise treatment selection.

\noindent
Recent work has shown that computational analysis of routine H\&E images can predict complementary pathology modalities, including IHC~\cite{latonen2024virtual, bai2023deeplearning},  mIF~\cite{wu2025rosie, valanarasu2025multimodal, li2026ai}, and spatially resolved transcriptomics~\cite{hoang2024deep, fu2025spatial}.  These results establish that tissue morphology encodes latent molecular information recoverable through learned mappings. Extracting this information accurately, however, remains a formidable challenge. Existing approaches rely on specialized generative models, each trained on a narrow paired dataset for a single source–target combination~\cite{yellapragada2024pathldm, zeng2022hist2st, chen2025visual, chelebian2025combining, liu2026leveraging}. This paradigm produces an ecosystem of siloed models that cannot share learned representations across modalities. Consequently, each new staining protocol, marker panel, or tissue type demands full retraining on dedicated paired data — a requirement that scales poorly and fundamentally limits clinical deployment~\cite{xu2025advancing, zhang2025content}.

\noindent
Addressing this limitation requires a unified generative framework capable of embedding heterogeneous pathology modalities within a common representational space, such that knowledge learned from one source–target pair transfers to others. A model with this property would enable virtual staining from routine histology, synthetic data augmentation across tissue types, and cross-modal imputation when specific assays are unavailable — all without requiring dedicated retraining for each task. Diffusion-based generative models offer a natural foundation for this goal. They have demonstrated state-of-the-art performance in high-fidelity synthesis across continuous domains~\cite{ho2020ddpm, rombach2022ldm, ma2024sit} and, critically, support flexible conditioning through shared denoising objectives in a common latent space~\cite{rombach2022ldm}, making them well suited to the heterogeneous, partially observed data characteristic of clinical pathology workflows.

\noindent
We introduce \ours (Multimodal Pathology Diffusion), a unified generative foundation model for histopathology that establishes H\&E-stained histology as a central bridging modality. By anchoring all generation tasks to the morphological richness of routine H\&E, \ours enables coherent cross-modal synthesis spanning transcriptomics, proteomics (e.g., IHC and mIF), tissue preparation protocols, and clinical text (Fig.~\ref{fig:fig1-Overview}a).

\noindent
\ours is built on a Diffusion Transformer\cite{peebles2023scalable} pretrained on a large-scale, multimodal corpus: 100 million H\&E image patches, 1.6 million text–histology pairs, and 10.8 million RNA–histology pairs drawn from 34 human organs (Fig.~\ref{fig:fig1-Overview}b). To handle heterogeneous conditioning inputs, which may be partially available at inference time, we introduce a decoupled cross-modal attention mechanism that processes text, RNA, and image embeddings through separate attention streams, avoiding interference between semantically distinct modalities (Fig.~\ref{fig:fig1-Overview}c).  

\noindent
The resulting model supports a broad range of cross-modal synthesis tasks with minimal or no task-specific fine-tuning. These include text-conditioned and RNA-conditioned histology generation for spatial transcriptomics exploration; translation of fresh-frozen (FF) sections to formalin-fixed paraffin-embedded (FFPE) appearance; and virtual staining of H\&E images into IHC and mIF assays. Across all evaluated benchmarks, \ours substantially outperforms domain-specific generative models trained for individual tasks (Fig.~\ref{fig:fig1-Overview}d), a result we attribute to the cross-modal priors acquired through large-scale multimodal pretraining.


\begin{figure*}[t]
    \centering
    \includegraphics[width=0.95\linewidth]{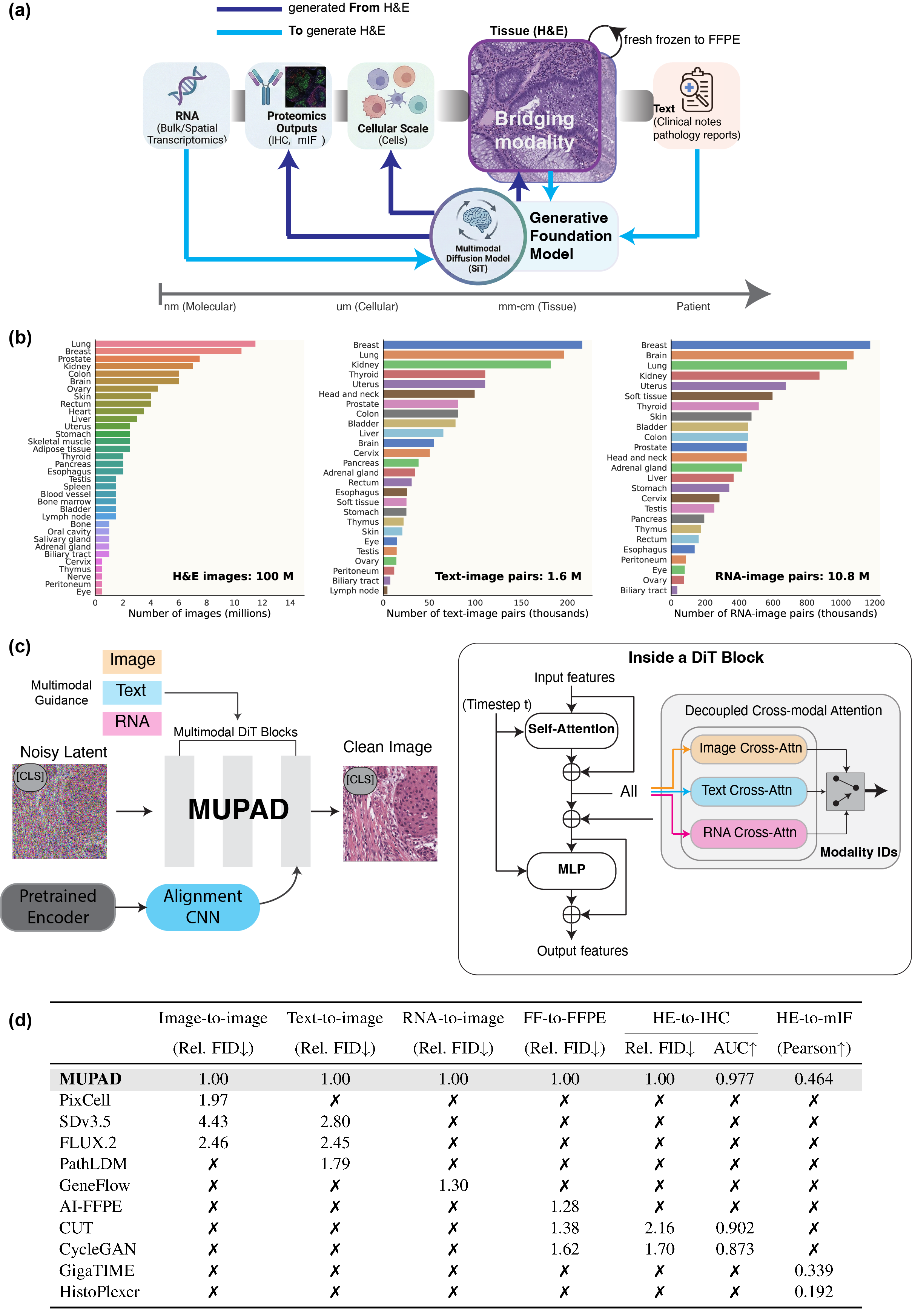}
\end{figure*}
\clearpage
\begin{figure*}[!ht]
    \caption{
        \textbf{Study Overview.}
        \textbf{a}, \ours framework. H\&E-stained histology serves as a bridging modality that integrates multi-scale data, from molecular transcriptomics and proteomics to tissue architecture and clinical text, enabling cross-modal generation across physical scales.
        \textbf{b}, Pretraining dataset, spanning 34 organs and comprising 100 million H\&E image patches, 1.6 million text-image pairs, and 10.8 million RNA--image pairs.
        \textbf{c}, The architecture of \ours employs DiT modules with decoupled cross-modal attention, which process image, text, and RNA conditioning signals in parallel streams. We use MUSK~\cite{xiang2025vision} to provide text/image guidance and Virchow2~\cite{zimmermann2024virchow2} for feature distillation during pretraining. 
        \textbf{d}, Benchmarking of \ours against other methods. We report Relative FID (normalized to \ours = 1.00) where lower is better; AUC for HE2IHC and Pearson Correlation for HE2mIF (higher is better for both).
    }
    \label{fig:fig1-Overview}
\end{figure*}

%% file: sections/2-results.tex
\heading{Image-to-image generation}

\begin{figure*}[!ht]
\centering
\includegraphics[width=0.86\linewidth]{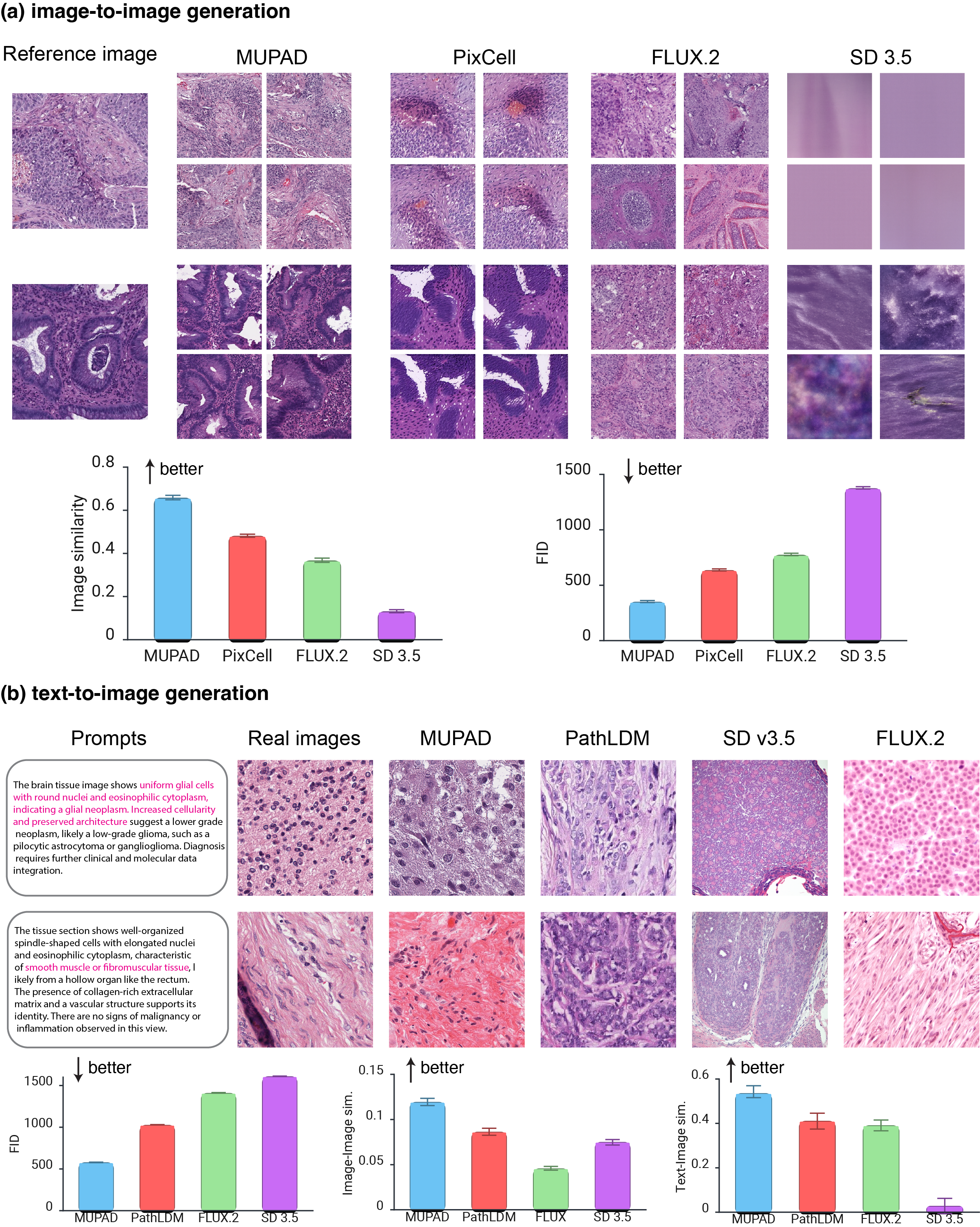}
\caption{
    \textbf{Image generation conditioned with image or text prompts.}
    \textbf{a}, Image-to-image generation.  {Representative examples and quantitative benchmarks demonstrate that \ours\ preserves authentic biological structures with greater fidelity than competing baselines, achieving superior Image--Image similarity and FID.}
    \textbf{b}, Text-to-image generation.  {Visual examples illustrate that \ours\ accurately reconstructs fine-grained histological features from text prompts. \ours\ achieves superior performance across Image--Image similarity, Text--Image alignment, and FID.}
     {All metrics are presented as median values with 95\% bootstrap confidence intervals. $\uparrow$ higher is better; $\downarrow$ lower is better.}
}
\label{fig:fig2-text2image-image2image}
\end{figure*}

\noindent
 {Image-to-image generation in pathology aims to synthesize realistic morphological variants from a reference tissue image while rigorously preserving the underlying histological architecture.}
We evaluated the performance of \ours\ against three baseline models: PixCell \cite{yellapragada2025pixcell}, Stable Diffusion 3.5 (SD v3.5) \cite{esser2024scaling}, and FLUX.2 \cite{flux-2-2025}.
 {Qualitative examples (Fig.~\ref{fig:fig2-text2image-image2image}a) reveal distinct performance differences across methods.}
\ours\ consistently generates images with high fidelity to the ground truth, preserving complex biological structures such as cellular density (Row 1) and glandular formations (Row 2).
In contrast, PixCell frequently introduces severe saturation artifacts and contrast distortions, often resulting in  {over-exposed} appearances with loss of cellular detail.
FLUX.2 attempts to mimic the histological texture but often produces stylized or grainy outputs that lack fine-grained structural coherence.
SD v3.5,  {a general-purpose natural image model without domain adaptation,} fails completely, generating abstract, smooth purple gradients devoid of any recognizable biological structures.
 {Together, these comparisons highlight that \ours\ maintains the morphological integrity.}

\noindent
\ours\ significantly outperforms all baselines across metrics for fidelity and  quality (Fig.~\ref{fig:fig2-text2image-image2image}a).
In terms of Image--Image Similarity, \ours\ achieves a score of 0.63 (95\% CI: [0.62, 0.64]), significantly surpassing PixCell (0.46, 95\% CI: [0.45, 0.47]), FLUX.2 (0.35, 95\% CI: [0.34, 0.36]), and SD v3.5 (0.12, 95\% CI: [0.11, 0.13]).
 {Regarding distributional quality, as measured by FID (lower values indicating closer alignment with the true image distribution),} \ours\ achieves a  {substantially lower} FID of 305.12 (95\% CI: [295.50, 315.80]) {, representing a 49.4\% improvement over PixCell (602.45, 95\% CI: [590.20, 615.10]), a 59.3\% improvement over FLUX.2 (750.30, 95\% CI: [740.10, 760.50]), and more than a fourfold improvement over SD v3.5 (1350.80, 95\% CI: [1340.20, 1360.50]).}

\heading{Text-to-image generation}

\noindent
{Text-to-image synthesis in pathology requires not only photorealistic rendering but also accurate instantiation of the specific biological entities described in natural language prompts.}
We benchmarked \ours\ against three baseline models: PathLDM \cite{yellapragada2024pathldm}, Stable Diffusion 3.5 (SD v3.5) \cite{esser2024scaling}, and FLUX.2 \cite{flux-2-2025}.
Visual examples (Fig.~\ref{fig:fig2-text2image-image2image}b) demonstrate that \ours\ captures fine-grained histological details that baseline models frequently miss.
For instance, when prompted to generate a lower-grade neoplasm characterised by ``uniform glial cells with round nuclei and eosinophilic cytoplasm,'' \ours\ correctly renders distinct, round nuclei within an appropriate eosinophilic background (Row 1).
In contrast, PathLDM produces distorted, elongated cellular structures that fail to capture the required uniform, round morphology.
Meanwhile, SD v3.5 fails to generate the correct cellular scale entirely, and FLUX.2 generates an overly synthetic, highly repetitive grid of cells.
Similarly, in the prompt describing ``smooth muscle'' (Row 2), \ours\ accurately synthesises well-organised spindle-shaped cells with elongated nuclei aligned along the muscle fibres.
Conversely, PathLDM hallucinates enlarged, round nuclei typical of malignant atypia rather than the requested benign tissue, and FLUX.2 generates hyper-intense, stylised staining that deviates from standard H\&E appearance.
 {These qualitative observations confirm that \ours\ has internalised the semantic correspondence between natural language descriptors and their histological manifestations.}

\noindent
\ours\ significantly outperforms all baselines across metrics for fidelity,  {text--image} alignment, and distributional quality (Fig.~\ref{fig:fig2-text2image-image2image}b).
 {In the FID evaluation, \ours\ achieves a score of 576.30 (95\% CI: [564.62, 587.98]), representing a 44.0\% improvement over PathLDM (1029.62, 95\% CI: [1020.13, 1039.10]), a 59.2\% improvement over FLUX.2 (1412.32, 95\% CI: [1402.78, 1421.87]), and a 64.2\% improvement over SD v3.5 (1610.85, 95\% CI: [1602.98, 1618.72]). Notably, PathLDM achieves a moderately lower FID than FLUX.2, suggesting some benefit from pathology-specific pretraining, yet both remain far behind \ours.}
In terms of Image--Image Similarity, \ours\ achieves a score of 0.11 (95\% CI: [0.10, 0.12]), significantly surpassing PathLDM (0.08, 95\% CI: [0.07, 0.09]), SD v3.5 (0.07, 95\% CI: [0.06, 0.08]), and FLUX.2 (0.04, 95\% CI: [0.03, 0.05]).
For Text--Image Alignment, \ours\  {secures the highest} adherence score of 0.53 (95\% CI: [0.50, 0.56]), followed by PathLDM (0.41, 95\% CI: [0.38, 0.44]) and FLUX.2 (0.39, 95\% CI: [0.36, 0.41]), with SD v3.5 showing negligible alignment (0.02, 95\% CI: [0.00, 0.05]).

\heading{Synthetic data generation for training augmentation}

\begin{figure*}[!t]
\centering
\includegraphics[width=0.95\linewidth]{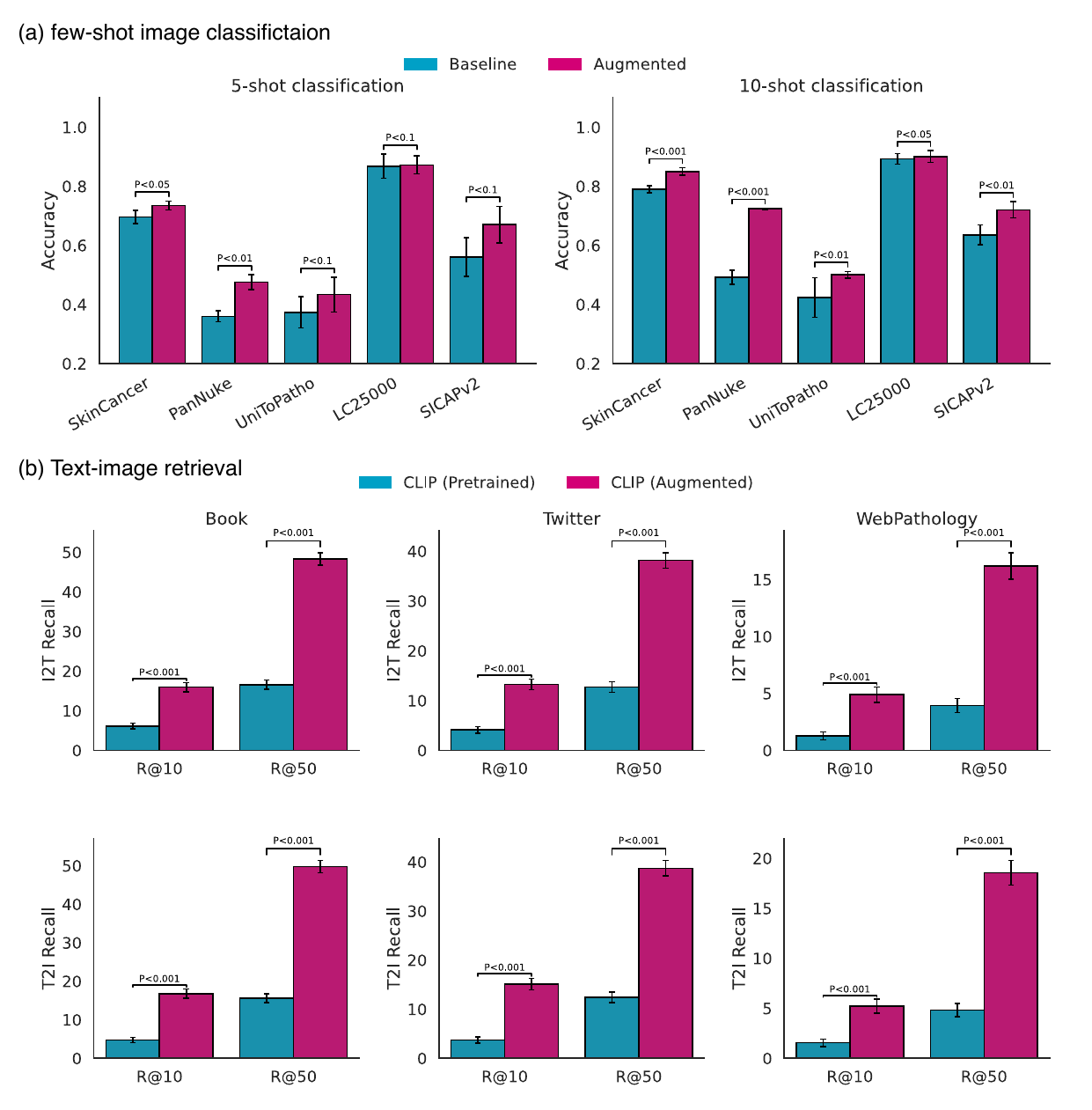}
\caption{\textbf{Training data augmentation using \ours.}
\textbf{a}, Few-shot classification augmented with \ours\ via image-to-image generation. Augmenting with \ours-synthesised morphological variants consistently improves classification accuracy across both 5-shot and 10-shot settings  {on} five evaluated datasets, demonstrating  {robust} generalisation under data-scarce conditions.
\textbf{b}, Pathology text--image retrieval augmented with \ours-generated image--text pairs. Fine-tuning a vanilla CLIP model on these synthetic pairs substantially improves both text-to-image (T2I) and image-to-text (I2T) retrieval at R@10 and R@50.}
\label{fig:fig4-downstream-augment}
\end{figure*}

\noindent
 {A key downstream application of high-fidelity generative models is data augmentation in label-scarce regimes. We therefore investigated whether synthetic data from \ours\ can serve as an effective augmentation strategy for } few-shot image classification and pathology-specific image--text retrieval (Fig.~\ref{fig:fig4-downstream-augment}).

\noindent
\textbf{Few-shot image classification.}
Few-shot classification aims to learn effective visual representations from only a handful of annotated examples (here, 5 or 10 per class).
We employed \ours\ to synthesise image variations conditioned on the limited real samples, thereby expanding each training set with plausible, class-consistent images.
Models were evaluated using a ResNet image encoder across five datasets.
Augmentation with \ours\ yielded consistent accuracy gains across all five datasets in both the 5-shot and 10-shot settings.
 {Improvements were particularly pronounced in the 10-shot regime:} PanNuke accuracy rose from 0.492 to 0.724, SkinCancer from 0.790 to 0.850, Unitopatho from 0.423 to 0.500, SICAPv2 from 0.635 to 0.720, and LC25000 from 0.892 to 0.900.
Analogous improvements were observed in the 5-shot setting, where PanNuke improved from 0.360 to 0.476 and SkinCancer from 0.696 to 0.735.
 {The magnitude of improvement was greatest for PanNuke with $+$47.2\% in 10-shot, underscoring the value of morphologically coherent synthetic augmentation when real training data are extremely scarce.}
These results confirm that \ours-synthesised images meaningfully improve feature representation in data-scarce pathology classification scenarios.

\noindent
\textbf{Pathology text--image retrieval.}
We next examined whether synthetic image--text pairs from \ours\ could strengthen vision--language alignment in the pathology domain.
Using \ours\ in text-to-image generation mode, we synthesised 100k pathology images paired with their conditioning text captions, and used these to fine-tune a vanilla CLIP model pre-trained on natural image--text pairs.
We evaluated retrieval performance at R@10 and R@50 across three independent pathology datasets (Book, Twitter, and WebPathology), covering both text-to-image (T2I) and image-to-text (I2T) directions.
Fine-tuning on \ours-augmented data produced substantial and consistent improvements across all datasets and retrieval directions.
On the Book dataset, I2T R@10 improved from 6.14\% to 15.94\% and T2I R@10 from 4.77\% to 16.76\%; at R@50, gains were similarly striking (16.55\%~$\to$~48.28\% for I2T; 15.56\%~$\to$~49.71\% for T2I).
On the Twitter dataset, I2T R@10 rose from 4.10\% to 13.21\% and T2I R@10 from 3.71\% to 15.10\%.
On WebPathology, the more challenging low-recall regime also benefited, with I2T R@10 increasing from 1.26\% to 4.88\% and T2I R@10 from 1.54\% to 5.23\%.
 {Across datasets, the consistent $\sim$3-fold gains in R@10 and $\sim$3-fold gains in R@50 demonstrate that \ours-generated image--text pairs effectively bridge the domain gap between general-purpose and pathology-specific domains, without requiring any manually curated paired data.}

\heading{Spatial transcriptomics to H\&E image generation}

\noindent
Understanding how transcriptional programs give rise to tissue morphology is a fundamental question in systems biology~\cite{coleman2024unlocking, chelebian2025combining, hieromnimon2025building}. Generative modelling of H\&E images conditioned on spatial transcriptomics (ST) profiles provides a direct computational approach, enabling systematic interrogation of the morphological determinants encoded within local gene expression~\cite{howard2024generative, wang2025geneflow}. 
To address this, we evaluated \ours\ on the task of conditionally generating H\&E images directly from ST profiles. Although \ours\ is pretrained on bulk RNA-seq, the transition to ST evaluation is seamless — both data sources are compressed into pathway-level scores using the same gene-set signatures, presenting the model with a consistent conditioning format across both stages. We assessed image fidelity and biological plausibility across five diverse human cancer datasets from MOSAIC\footnote{\url{https://ega-archive.org/dacs/EGAC50000000398}}: bladder cancer, glioblastoma (GBM), diffuse large B-cell lymphoma (DLBCL), mesothelioma, and ovarian cancer, encompassing a total of 54,033 H\&E--ST Visium spots.

\begin{figure*}[!ht]
\centering
\includegraphics[width=\linewidth]{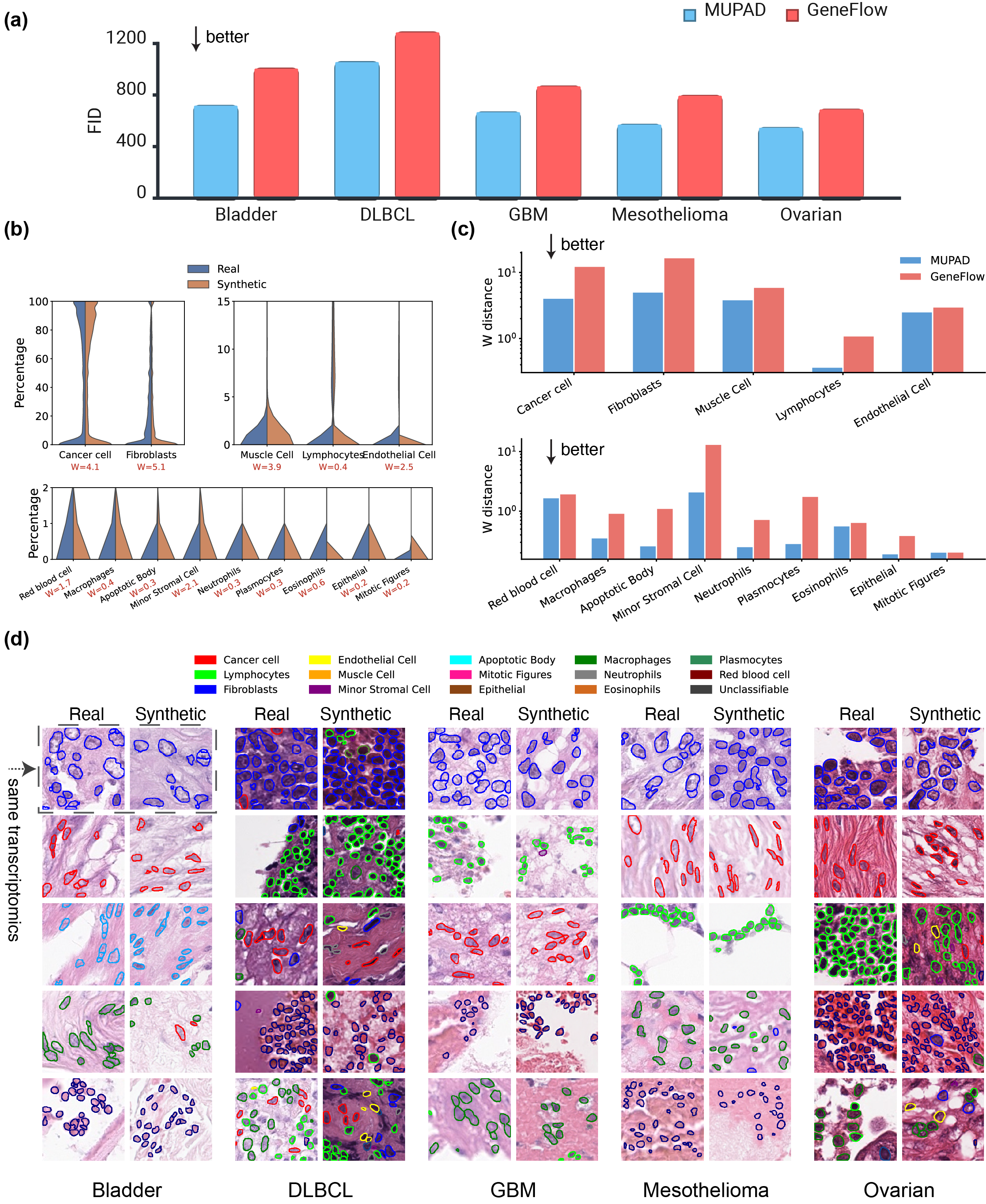}
\end{figure*}
\clearpage
\begin{figure*}
\caption{
\textbf{H\&E image generation from spatial transcriptomics using \ours.}
\textbf{a}, FID of synthetic H\&E images across five cancer types. Lower FID indicates greater fidelity to real images paired with the same transcriptomic profiles.
\textbf{b}, Cell-type composition in synthetic H\&E images (bladder cancer). Violin plots compare the relative abundance of 14 cell types between real (blue) and synthetic (orange) images from \ours\ and GeneFlow, estimated using Histoplus~\cite{histoplus2024}. The Wasserstein distance ($W$) quantifies distributional alignment of cell composition between synthetic and ground-truth images {; lower values indicate closer agreement.}
\textbf{c}, Wasserstein distance comparison between \ours\ and GeneFlow for bladder cancer. Pan-cancer results in Extended Data Fig.~\ref{fig:extended_st2image}.
\textbf{d}, Representative real and synthetic H\&E image pairs from \ours\ conditioned on matched spatial transcriptomics.
}
\label{fig:fig5-st2image}
\end{figure*}

\noindent
 {We first compared distributional image fidelity between \ours\ and} GeneFlow~\cite{wang2025geneflow}  {using FID (Fig.~\ref{fig:fig5-st2image}a). \ours\ consistently produced synthetic images with substantially higher visual and statistical fidelity to the true tissue morphology across all five evaluated cancer types.}
In bladder cancer, \ours\ achieved an FID of 724.6, a 28.4\% improvement over GeneFlow (1012.5).
Compelling performance margins were consistently observed in mesothelioma (577.2 vs.\ 801.4), ovarian cancer (553.1 vs.\ 695.4), GBM (674.2 vs.\ 874.2), and DLBCL (1063.0 vs.\ 1293.5).

\noindent
Beyond image quality, we investigated whether the generative models accurately recapitulated the cellular compositions dictated by the input transcriptomic profiles.
We employed Histoplus~\cite{histoplus2024} to perform downstream cell-type composition analysis by classifying morphological structures into 14 distinct cell types, including cancer cells, fibroblasts, lymphocytes, and macrophages.
 {We then quantified the distributional alignment of inferred cell-type proportions between each set of synthetic images and the real H\&E images paired with identical transcriptomic inputs (Fig.~\ref{fig:fig5-st2image}b, Extended Data Fig.~\ref{fig:extended_st2image}).}
\ours\ faithfully preserved the intricate cell-type distributions observed in real tissue, significantly outperforming GeneFlow.
Violin plots comparing the relative abundances of key cellular populations demonstrate well-aligned distributions between real and \ours-generated images (Fig.~\ref{fig:fig5-st2image}b), quantitatively supported by markedly lower Wasserstein distances across nearly all detected cell types (Fig.~\ref{fig:fig5-st2image}c).
 {Visual inspection of segmented image tiles confirms that \ours\ not only captures broad compositional proportions but also reconstructs realistic cellular morphologies and spatial arrangements conditioned on the transcriptomic input (Fig.~\ref{fig:fig5-st2image}d); for example, immune and stromal gene signatures appropriately yielded synthetic morphological regions enriched with the corresponding cell types.}

\heading{Fresh frozen to FFPE slide image translation}

\noindent
 {Intra-operative assessment of tissue biopsies relies on fresh frozen (FF) sections, which enable rapid turnaround but introduce cryopreservation artefacts, including ice crystal damage, tissue folds, and compromised nuclear detail, that limit diagnostic accuracy~\cite{yuan2025ai}.}
The gold standard formalin-fixed paraffin-embedded (FFPE) processing yields superior morphological preservation, yet requires 12--48 hours, rendering it incompatible with surgical timelines~\cite{borah2023rapid, ozyoruk2022deep}.
Here, we evaluated \ours\ for virtual FF-to-FFPE conversion {, a task that could substantially reduce intra-operative turnaround time while preserving diagnostic quality.}
We benchmarked against AI-FFPE~\cite{ozyoruk2022deep}, CUT~\cite{park2020contrastive}, and CycleGAN~\cite{zhu2017unpaired} using \textbf{lung} and \textbf{brain} tissue samples from TCGA (Extended Data Fig.~\ref{fig:fig4-ff2ffpe}).
\ours\ demonstrates robust improvements in fidelity as measured by FID.
For \textbf{lung tissue}, \ours\ achieves the lowest FID of 323.7 (95\% CI: [319.6, 327.7]), substantially outperforming AI-FFPE (435.3, 95\% CI: [430.8, 439.8]), CUT (495.0, 95\% CI: [490.2, 499.8]), and CycleGAN (508.8, 95\% CI: [503.7, 513.9]) {, corresponding to relative FID reductions of 25.6\%, 34.6\%, and 36.4\%, respectively.}
In the \textbf{brain tissue} cohort, \ours\ maintains this advantage with a FID of 589.3 (95\% CI: [581.4, 597.2]), substantially lower than AI-FFPE (731.0, 95\% CI: [723.5, 738.4]), CUT (765.7, 95\% CI: [757.8, 773.6]), and CycleGAN (968.2, 95\% CI: [959.5, 976.9]).

\heading{H\&E to IHC image translation}

\begin{figure*}[!ht]
\centering
\includegraphics[width=0.91\linewidth]{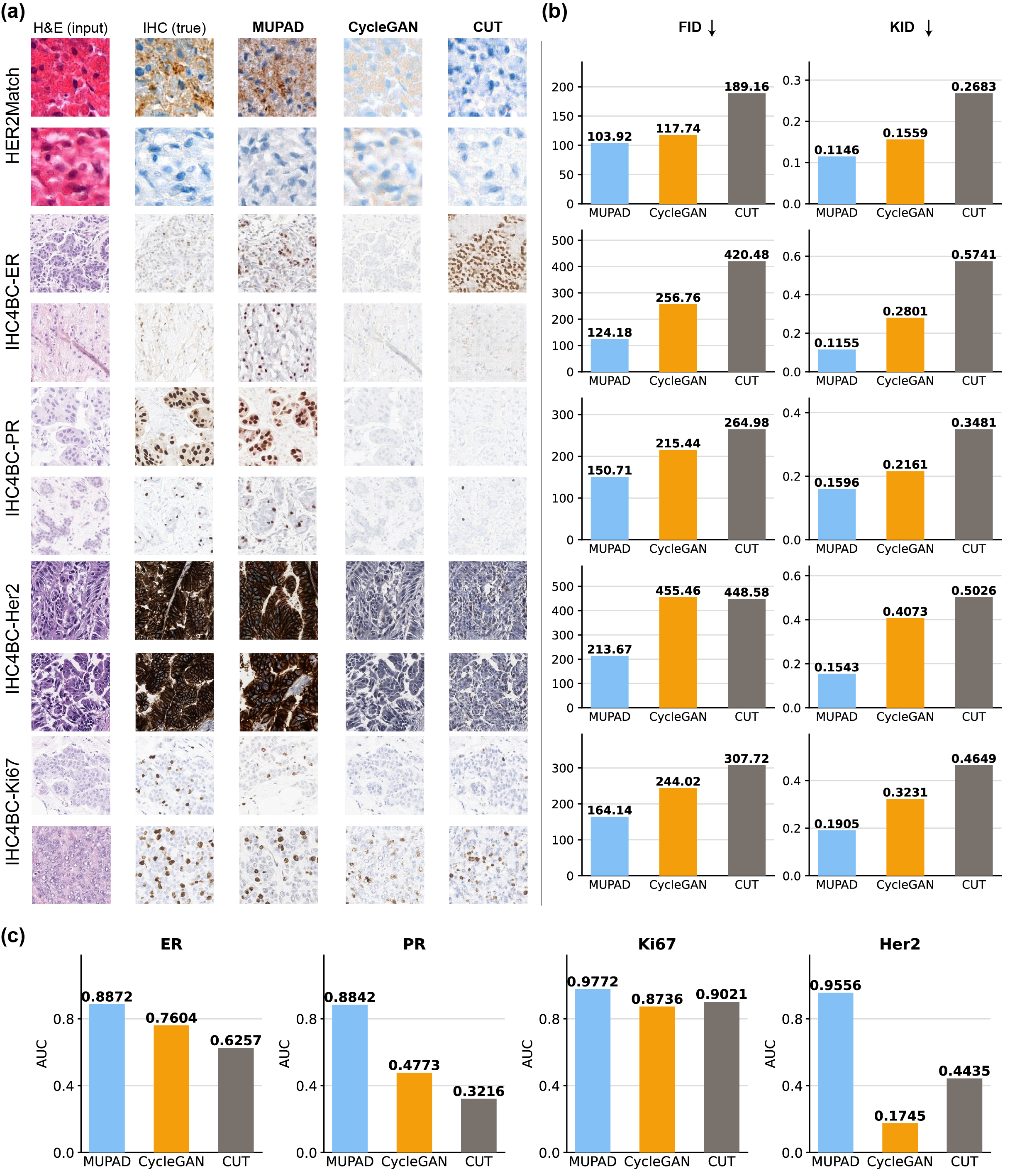}
\caption{
    \textbf{Virtual H\&E-to-IHC translation and clinical validation.}
    \textbf{a,} Visual examples of multi-stain virtual IHC generation. Compared to CycleGAN and CUT, \ours\ provides more accurate spatial stain rendering.
    \textbf{b,} FID and KID scores demonstrate that \ours\ achieves better distributional fidelity and perceptual quality.
    \textbf{c,} Clinical utility on the IHC4BC dataset. When using virtual IHC images to predict ground-truth clinical biomarkers (H-scores for ER/PR/Ki67; amplification status for HER2), \ours\ yields the highest AUC performance across all markers.  {Sub-0.5 AUC values for baseline methods indicate systematic \textit{inverse} correlations with clinical labels, as observed the semantic misalignment in stain generation.}
}
\label{fig:fig5-he2ihc}
\end{figure*}

\noindent
IHC staining remains indispensable for protein biomarker assessment and treatment stratification, but requires dedicated laboratory infrastructure, trained personnel, and multi-day turnaround times, limiting its availability in resource-constrained clinical settings.~\cite{klockner2025h}.
We  {evaluated} \ours\ on the task of virtual IHC staining from standard H\&E slides, focusing on five distinct markers in the \textit{weakly co-registered} IHC4BC~\cite{Akbarnejad2023PredictingKE} and HER2Match~\cite{HER2match_data} datasets.
We benchmarked against widely used image-to-image translation methods CUT~\cite{park2020contrastive} and CycleGAN~\cite{zhu2017unpaired}.

\noindent
Visual examples (Fig.~\ref{fig:fig5-he2ihc}a) demonstrate that \ours\ significantly mitigates the hallucinations and colour artifacts common in baseline methods.
For nuclear markers (IHC4BC-Ki67, IHC4BC-ER, IHC4BC-PR), \ours\ exhibits precise localisation of positive tumour nuclei while correctly suppressing background stromal signals.
In contrast, CycleGAN frequently generates washed-out staining with poor contrast, failing to distinguish between weak and strong positivity.
CUT conversely often introduces high-frequency noise and tiling artefacts, leading to granular, unrealistic textures and over-staining of the background.
For membrane markers (IHC4BC-Her2), \ours\ preserves the crisp, continuous membranous staining pattern required for accurate HER2 scoring.
Baselines struggle with this morphological constraint: CycleGAN produces patchy, discontinuous staining, while CUT exhibits significant structural distortion and unrealistic intensity gradients.

\noindent
\ours\ achieves state-of-the-art performance across all five datasets using FID and KID (lower scores indicating distributions closer to real clinical IHC images; Fig.~\ref{fig:fig5-he2ihc}b) {; a pathology foundation model~\cite{hoptimus1} was used as the encoder for both metrics.}
\ours\ consistently yields the lowest FID scores, often outperforming the nearest competitor by a wide margin.
For instance, on \textbf{IHC4BC-ER}, \ours\ achieves an FID of 124.18, a roughly 50\% reduction compared to CycleGAN (256.76) and a threefold improvement over CUT (420.48).
A similar trend is observed for \textbf{IHC4BC-Her2}, where \ours\ records an FID of 213.67 compared to 455.46 for CycleGAN and 448.58 for CUT.
The KID metric further substantiates this dominance, with \ours\ achieving the lowest kernel distances across all tasks (e.g., 0.1146 for HER2Match and 0.1155 for IHC4BC-ER compared to 0.2683 and 0.5741 for CUT, respectively).

\noindent
Beyond visual and statistical fidelity, we evaluated the  {downstream} \textbf{clinical utility} of virtual IHC on the IHC4BC dataset~\cite{Akbarnejad2023PredictingKE}.
 {Predictions from each model's virtual IHC images were evaluated against H-scores for ER, PR, and Ki67, and against binary HER2 amplification labels (negative/0--1+ vs.\ positive/3+; equivocal 2+ cases excluded).}
As shown in Fig.~\ref{fig:fig5-he2ihc}c, \ours\ demonstrates superior diagnostic reliability, achieving AUC scores that significantly exceed those of baseline models across all biomarkers.
Notably, for PR and HER2, \ours\ maintains high predictive accuracy (AUC~=~0.8842 and 0.9556, respectively), whereas CycleGAN and CUT performance drops precipitously (e.g., CUT reaching only 0.3216 for PR and CycleGAN falling to 0.1745 for HER2).
The sub-0.5 AUC values for these baselines indicate a systematic inverse correlation with clinical labels,  {most likely resulting from semantic misalignment in which these models consistently hallucinate staining in histologically negative regions or suppress signal in positive ones.}
For Ki67, \ours\ achieves near-perfect alignment with clinical ground-truth proliferation indices (AUC~=~0.9772).
 {Collectively, these results indicate that \ours-generated virtual IHC images are not only visually coherent but also statistically and diagnostically aligned with real clinical IHC slides, suggesting potential utility as a cost-effective surrogate for protein biomarker assessment.}

\heading{H\&E to mIF image translation}

\noindent
 {Multiplex immunofluorescence (mIF) enables simultaneous co-localised profiling of multiple protein markers on a single tissue section while preserving spatial architecture, providing a far richer characterisation of the tumour microenvironment than conventional IHC. Despite its promise, mIF remains technically demanding and cost-prohibitive at scale, limiting its widespread clinical adoption~\cite{liu2025spatial}.}
We evaluated the capacity of \ours\ to bridge the gap between standard histology and high-dimensional spatial biology by predicting mIF protein expression maps directly from H\&E images.
We utilised the ORION-CRC dataset~\cite{lin2023high} and benchmarked performance against two recent methods specialised for this task: GigaTIME~\cite{valanarasu2025multimodal} and HistoPlexer~\cite{andani2025histopathology}.
 {We report two complementary metrics: patch Pearson Correlation Coefficient (PCC), adopted from GigaTIME, which measures local spatial fidelity within individual patches and captures fine-grained morphological correspondence; and slide-level PCC, which aggregates patch-mean intensities across the whole slide to provide a more robust measure of global expression variation that is less sensitive to local noise (see Methods for details).}

\noindent
\ours\ establishes a new state of the art in virtual spatial proteomics, demonstrating significantly higher correlation with ground truth data across the marker panel (Fig.~\ref{fig:fig6-he2mIF}a,b).
 {Analysis of the \textbf{patch PCC} reveals that \ours\ significantly outperforms competing methods ($P \leq 0.001$, two-sided paired permutation test) on 12 out of 15 markers, achieving the highest average patch PCC of 0.238, compared to 0.198 for GigaTIME and 0.071 for HistoPlexer.}
Specifically, \ours\ achieves a patch PCC of 0.398 (95\% CI: [0.325, 0.468]) for E-cadherin and 0.336 (95\% CI: [0.277, 0.381]) for Pan-CK, representing a substantial improvement over GigaTIME (E-cadherin: 0.301; Pan-CK: 0.265) and HistoPlexer (E-cadherin: 0.098; Pan-CK: 0.091).
The performance gap is even more pronounced for difficult-to-predict immune markers; for PD-L1, \ours\ achieves a patch PCC of 0.218 (95\% CI: [0.147, 0.288]), outperforming GigaTIME (0.150) and far exceeding the near-zero correlation of HistoPlexer (0.004).

\noindent
At the \textbf{slide level}, \ours\ similarly leads across the panel, achieving the highest average slide-level PCC of 0.464, compared to 0.339 for GigaTIME and 0.192 for HistoPlexer, and ranking first on 11 out of 15 markers.
For structural markers, \ours\ attains slide-level PCCs of 0.590 (95\% CI: [0.524, 0.631]) for E-cadherin, substantially outperforming GigaTIME (0.437) and HistoPlexer (0.173); performance on Pan-CK is comparable between \ours\ (0.549) and GigaTIME (0.565).
For the clinically relevant immune marker PD-L1, \ours\ achieves a slide-level PCC of 0.220 (95\% CI: [0.106, 0.345]), outperforming GigaTIME (0.092),  {though HistoPlexer achieves a comparable slide-level score (0.241), likely reflecting the high inter-slide variability intrinsic to this sparsely expressed marker.}
 {The markers where competing methods maintain a modest advantage (Pan-CK, CD45RO, and Ki67 for GigaTIME) are characterised by high spatial variability or smooth expression gradients at the slide level, suggesting avenues for further architectural refinement.}

\noindent
Visual examples (Fig.~\ref{fig:fig6-he2mIF}c)  {further elucidate the model's capacity to infer complex molecular distributions solely from morphological images.}
\ours\ generates marker expression maps that faithfully recapitulate the ground-truth spatial heterogeneity.
For structural markers such as E-cadherin and Pan-CK, \ours\ accurately delineates the epithelial tumour architecture, preserving sharp tumour--stroma boundaries (Pan-CK patch PCC: 0.817).
In stark contrast, HistoPlexer struggles to localise these signals, often yielding inverted or noisy predictions (Pan-CK patch PCC: $-$0.177), while GigaTIME produces diffuse outputs that lack morphological precision.
Crucially, \ours\ demonstrates exceptional sensitivity in detecting sparse immune markers.
For PD-L1, a critical biomarker for immunotherapy response, \ours\ correctly identifies localised expression ``hotspots'' (PCC: 0.615), whereas GigaTIME (PCC: 0.215) and HistoPlexer (PCC: 0.112) fail to distinguish specific cellular positivity from background noise.
Similarly, for CD45 and CD68, \ours\ successfully reconstructs the density and distribution of immune infiltrates and significantly outperforms baselines.

\begin{figure*}[!t]
\centering
\includegraphics[width=0.9\linewidth]{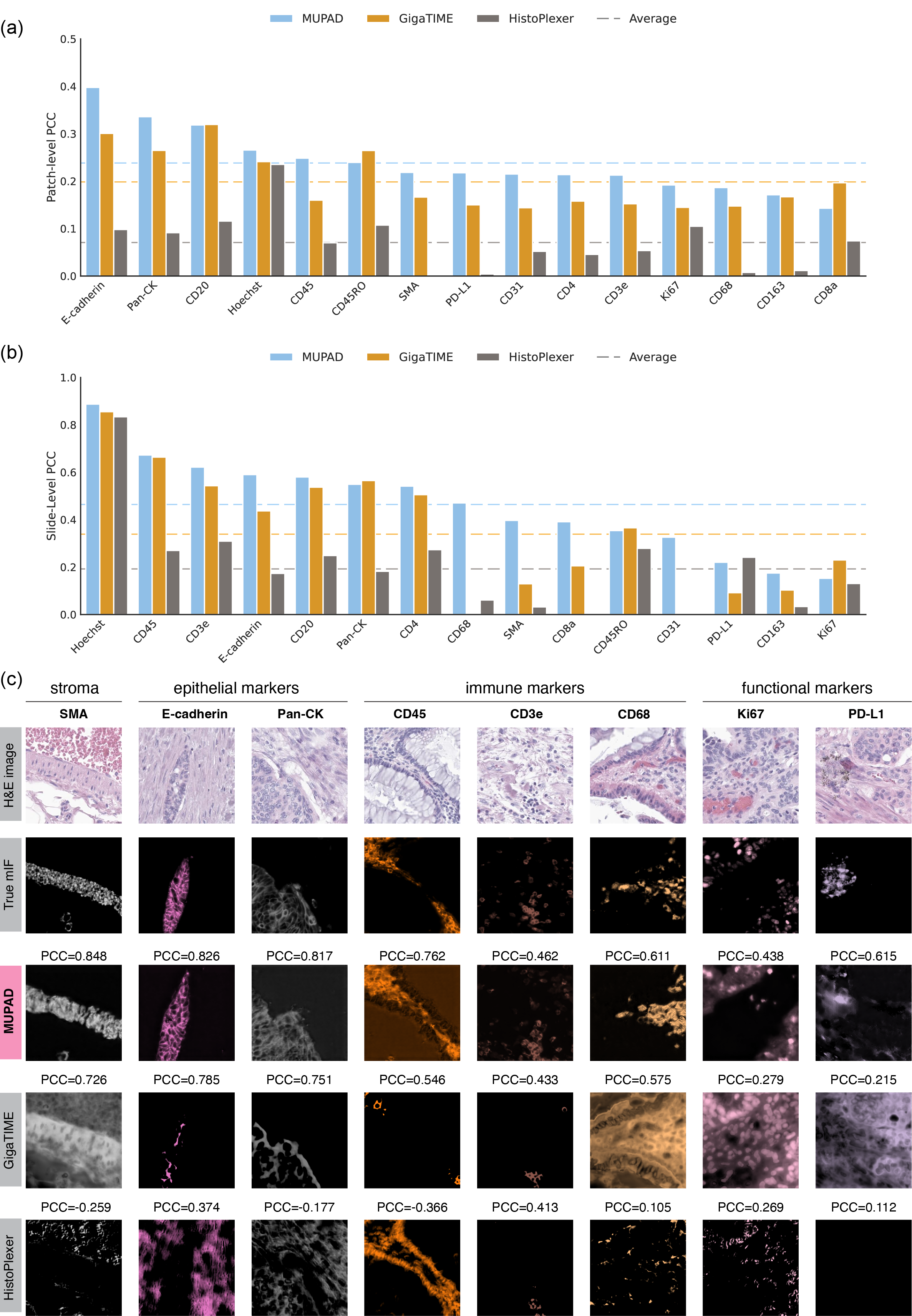}
\end{figure*}
\clearpage
\begin{figure*}
\caption{
    \textbf{H\&E to mIF image translation.}
    Quantitative comparison of marker-level prediction accuracy measured by patch Pearson Correlation Coefficient (PCC) (\textbf{a}) and slide-level PCC (\textbf{b}) across 15 protein markers. \ours\ achieves the highest average patch PCC (0.238) and slide-level PCC (0.464), consistently surpassing GigaTIME (0.198; 0.339) and HistoPlexer (0.071; 0.192) across both structural and immune markers. The dashed lines represent the average performance across all markers for each method.
    \textbf{c}, Representative visualisations of structural (Hoechst, E-cadherin, Pan-CK) and immune (CD45, CD3e, CD68, Ki67, PD-L1) markers. The top row displays the input H\&E image, followed by the ground truth mIF and the spatial expression maps predicted by each model. Pixel-wise PCC scores for the displayed regions are provided above each prediction, highlighting \ours's superior reconstruction of fine-grained morphological structures and sparse immune infiltrates compared to baselines.
     { Statistical significance was assessed using two-tailed paired permutation tests (${\ast}P \leq 0.05$; ${\ast\ast}P \leq 0.01$; ${\ast\ast\ast}P \leq 0.001$).}
}
\label{fig:fig6-he2mIF}
\end{figure*}


\heading{Ablation study of model design}

\noindent
To  {systematically} validate the  {key} design choices of \ours, we conducted ablation studies on: (1) the Decoupled Cross-Attention (DCA) mechanism for multimodal condition integration, and (2) the CNN-based spatial alignment loss with a CLS token denoising objective.
All ablations use a SiT-B/2 backbone and share identical training configurations to ensure fair comparison.

\noindent
\textbf{Decoupled Cross-Attention versus Shared Cross-Attention.}
A natural baseline for multimodal conditioning is to concatenate all condition embeddings---image, text, and transcriptomic (RNA)---and feed them into a \emph{shared} cross-attention module, augmented with modality embeddings to discriminate sources.
 {While conceptually straightforward, this design forces all modalities to compete within a common key--value projection space.}
The high-dimensional and statistically heterogeneous nature of RNA pathway embeddings, natural language tokens, and visual patch embeddings means that a shared projection cannot simultaneously capture the distinct structural priors of each modality.
 {As a result, gradient signals from different modalities interfere during training, degrading generation quality across all tasks.}

\begin{figure*}[h!]
    \centering
    \includegraphics[width=0.82\linewidth]{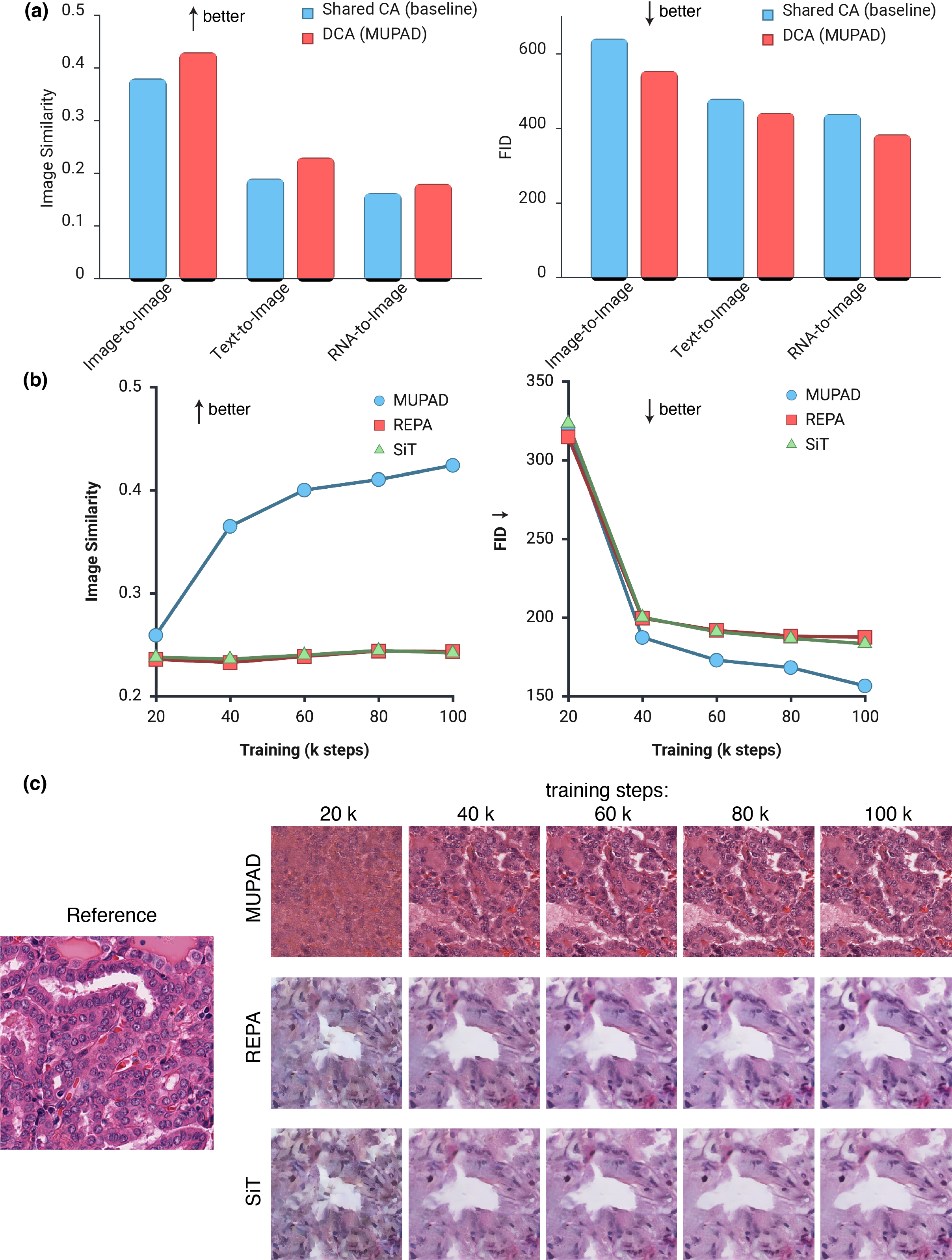}
    \caption{\textbf{Ablation studies of \ours.}
    \textbf{(a)} Comparison of the proposed decoupled cross-attention (DCA) against shared cross-attention across multimodal embeddings. DCA consistently improves performance across image-to-image, text-to-image, and RNA-to-image generation tasks {, with relative FID reductions of 13.6\%, 7.9\%, and 12.4\%, respectively.}
    \textbf{(b)} FID and image similarity  {trajectories} for image-to-image generation of SiT, REPA, and \ours. Lower FID and higher image similarity indicate better performance.
    \textbf{(c)} Visual comparison of image-to-image generation across training steps (20K--100K) for \ours, REPA~\cite{yu2025representation}, and SiT~\cite{ma2024sit}.
    }
    \label{fig:fig-ablation}
\end{figure*}

\noindent
Our DCA module instead processes each modality through its own dedicated, parallel attention stream (see Methods), fusing modality-specific contributions additively before the subsequent feed-forward block.
To quantify this difference, we trained both variants and evaluated FID and image similarity across three generation tasks: image-to-image, text-to-image, and RNA-to-image.
As shown in Fig.~\ref{fig:fig-ablation}a, replacing DCA with shared cross-attention consistently and substantially degrades performance across all tasks and metrics.
 {For FID, the shared cross-attention variant scores 641.72, 480.12, and 439.01 on image-to-image, text-to-image, and RNA-to-image, respectively, compared to 554.43, 442.31, and 384.54 for \ours, representing relative FID reductions of 13.6\%, 7.9\%, and 12.4\%.}
The degradation is particularly pronounced for image-to-image generation, where the visual conditioning signal is likely suppressed by the noisier RNA and text modalities under shared attention.
Cosine similarity scores further verify this trend: \ours\ achieves 0.43, 0.23, and 0.18 across the three tasks, consistently outperforming the shared cross-attention baseline (0.38, 0.19, and 0.162).
These results confirm that modality-specific attention pathways are critical for maintaining the fidelity and semantic coherence of each conditioning modality within a unified generative framework.

\noindent
\textbf{Alignment Loss Design.}
Beyond the attention architecture, we investigated the effect of the spatial alignment and semantic regularisation objectives used during \ours\ pretraining.
We compared three variants:
(i) \textbf{Naive denoising loss} (SiT~\cite{ma2024sit}): the standard velocity-prediction flow-matching objective with no additional alignment term;
(ii) \textbf{REPA}~\cite{yu2025representation}: an MLP-based alignment strategy that projects intermediate denoiser representations onto a pretrained encoder~\cite{zimmermann2024virchow2} feature space via a lightweight multi-layer perceptron;
(iii) \textbf{\ours\ alignment}: our proposed CNN-based spatial alignment with an auxiliary \texttt{CLS} token denoising loss that enforces global semantic consistency between noisy intermediate representations and the clean image's semantic embedding throughout the denoising trajectory.
 {The rationale for \ours's design is twofold: the CNN spatial alignment preserves the fine-grained histological texture information critical for high-fidelity pathology image synthesis, while the \texttt{CLS} token denoising loss anchors the global semantic content of intermediate noisy representations to the clean-image manifold.}

\noindent
We evaluated convergence on the image-to-image generation task, plotting image similarity and FID against training steps in Fig.~\ref{fig:fig-ablation}b.
 {\ours\ improves steadily and monotonically from an image similarity of 0.259 at 20K steps to 0.424 at 100K steps---nearly twice the similarity achieved by both REPA (0.243) and SiT (0.242), which remain largely stagnant throughout training.}
FID scores are consistent with this trend: \ours\ reaches 156.39 at 100K steps, outperforming REPA (187.39) and SiT (183.22) by 16.5\% and 14.6\%, respectively.
Notably, REPA does not confer a meaningful advantage over the naive SiT baseline in our setting, suggesting that MLP-based global alignment is insufficient for capturing the fine-grained spatial structure of histological images.
We also provide visual examples illustrating model convergence at representative training checkpoints (Fig.~\ref{fig:fig-ablation}c).
 {Taken together, these results demonstrate that the combination of CNN spatial alignment and the \texttt{CLS} token denoising loss provides training signals that substantially accelerate convergence and improve both semantic fidelity and generation quality relative to existing alignment strategies.}

%% file: sections/3-discussion.tex
\noindent
Generative models in computational pathology have thus far been developed one task at a time~\cite{yellapragada2025pixcell, yellapragada2024pathldm,wang2025geneflow,valanarasu2025multimodal,andani2025histopathology}.  Each model is trained from scratch on a dedicated paired dataset and cannot transfer learned representations to related tasks. \ours departs from this paradigm. A single pretrained model, without task-specific retraining, simultaneously supports image-, text-, and RNA-conditioned generation, virtual staining of IHC and multiplex immunofluorescence, and synthetic data augmentation for downstream classification. This breadth is not merely convenient. It reflects a deeper structural claim: superficially distinct generation tasks in pathology are projections of a shared cross-modal biological distribution, and a model trained at sufficient scale can learn to represent that distribution directly.  Drawing on the success of large-scale foundation models~\cite{lu2024visual, chen2024towards, xiang2025vision, chen2024titan, vorontsov2024foundation}, we argue that the field stands to gain more from scaling a unified generative model that encodes the joint distribution across histology, molecular profiles, and clinical language than from continuing to expand an ecosystem of narrow, non-transferable specialist models.

\noindent
The biological basis for this unification lies in the role of H\&E histology as a natural cross-modal bridge. Tissue morphology is not independent of molecular state, but rather an observable consequence of the underlying transcriptional programs and protein expression patterns~\cite{elmentaite2024profiling, valanarasu2025multimodal}.
Because H\&E histology integrates these effects into a single observable, a generative model conditioned on diverse molecular and textual inputs can, in principle, learn to invert this mapping — recovering latent molecular states from morphological appearance. \ours's ability to preserve cell-type composition during RNA-conditioned generation, localize immune markers from morphological inputs alone, and translate clinical text into biologically coherent tissue architectures provides empirical evidence that this inversion is achievable at scale.

\noindent
The virtual staining results bear on a fundamental question about the information content of routine H\&E. \ours predicts spatially resolved immune and stromal markers — including CD3, CD68, CD31, and SMA — from H\&E input alone, with substantially higher fidelity than specialist baselines\cite{valanarasu2025multimodal,andani2025histopathology}. A key reason for this performance is that \ours is trained to synthesize the full spatial distribution of marker expression rather than a reduced scalar summary, allowing it to exploit morphological correlates of protein localization at finer structural scales than discriminative models typically access. The FF-to-FFPE translation results point to a complementary finding: even structurally degraded cryo-sections retain sufficient morphological information for a generative model to recover diagnostically faithful tissue architecture, suggesting that the information loss introduced by freeze artifacts is largely recoverable when the model has learned the appropriate tissue priors.

\noindent
The training data augmentation results extend this argument beyond image quality.
Improvements in few-shot classification and cross-modal retrieval from \ours-generated samples indicate that the model's outputs carry transferable biological priors, not merely plausible visual textures.
\ours serve as a form of structured data synthesis that is genuinely complementary to supervised learning on real data.
The implications are most consequential in data-scarce settings of rare tumour subtypes, novel biomarker panels, and low-resource clinical environments~\cite{moor2023foundation,fountzilas2025convergence,yang2025multimodal}, where the acquisition of large annotated cohorts is impractical and where synthetic augmentation guided by biological prior knowledge could meaningfully reduce annotation burden.

\noindent
The ablation studies yield insights of \ours's architecture design.
The failure of shared cross-attention to match the performance of Decoupled Cross-Attention (DCA) illustrates a general principle: modalities as structurally distinct as natural language tokens, spatial gene expression vectors, and visual image patches cannot be effectively fused by forcing them to compete within a common key-value space.
Allocating each modality its own parallel attention stream before additive integration reflects a more fundamental requirement that is likely to generalise to other multimodal generative settings~\cite{rombach2022ldm}.
Equally, the superiority of CNN-based spatial alignment over the MLP-based global alignment of REPA~\cite{yu2025representation} reflects a domain-specific consideration: pathology images carry diagnostically meaningful structure simultaneously across spatial scales, from nuclear morphology to glandular architecture, and preserving this hierarchy during the denoising trajectory requires spatially aware objectives rather than globally aggregated losses.

\noindent
Several limitations of this work point to clear directions for future development. Diffusion-based generation is inherently stochastic, and it is possible for \ours to produce anatomically implausible outputs under distribution shift. Integrating pathology-specific discriminative classifiers as post-hoc quality filters is one natural safeguard, though a more comprehensive solution will require formal uncertainty quantification within the generative process itself. More critically, clinical deployment of virtual staining demands prospective validation of concordance between computationally predicted biomarker scores and laboratory-measured ground truth. Until such validation is performed across representative patient populations, \ours's virtual staining outputs should be regarded as hypothesis-generating rather than diagnostic.

\noindent
In conclusion, by explicitly modelling the joint distribution across histological, molecular, and clinical modalities, \ours advances understanding of the information content encoded in routine histology, expands the potential for diagnostic access in settings where specialized assays are unavailable, and provides a scalable substrate for training future discriminative models on synthetic multimodal data. As pretraining corpora grow in scale and additional modalities such as spatial proteomics are incorporated, unified generative frameworks of this kind have the potential to become core infrastructure for computational pathology.

%% file: sections/4-methods.tex
\heading{Pretraining Datasets}

\noindent
We curated a large-scale multimodal corpus spanning morphological, molecular, and textual data. The dataset comprises three components: (1)~\textbf{H\&E images} --- approximately 100 million $512{\times}512$ patches at 20$\times$ magnification from TCGA\cite{Weinsteinetal2013}, GTEx\cite{Lonsdaleetal2013}, PAIP\cite{kim2023paip}, and PLCO\cite{Zhuetal2013}; (2)~\textbf{text--image pairs} --- 1.6 million descriptive histopathological caption-image pairs from PathGen\cite{sun2024pathgen}; and (3)~\textbf{RNA--image pairs} --- 10.8 million bulk RNA-seq profiles from TCGA linked to their matched WSIs. Raw RNA-seq counts were normalized to TPM, filtered to remove housekeeping and mitochondrial genes, and compressed into 331 pathway-level enrichment scores using established pan-cancer gene signatures\cite{jaume2024modeling}, yielding biologically interpretable conditioning inputs.

\heading{Model Architecture and Pretraining}

\noindent
We propose a unified multimodal Scalable Interpolant Transformer~\cite{ma2024sit} (SiT) that jointly conditions image generation on transcriptomics, text, and reference images.
Let $\mathbf{x}_0$ be the clean H\&E image and $\mathbf{x}_t$ its noisy version at generative step $t$. The model predicts $\mathbf{x}_0$ conditioned on $\mathcal{C} = \{\mathbf{c}_{\text{RNA}}, \mathbf{c}_{\text{text}}, \mathbf{c}_{\text{image}}\}$.

\noindent
\textbf{Decoupled Cross-Attention (DCA).}
Directly fusing text, RNA, and image features hurts performance because these modalities carry information at very different scales and semantics. Our DCA module processes each modality in independent parallel attention streams, all queried by the intermediate image representation $\mathbf{h}$:
\begin{equation}
    \text{DCA}(\mathbf{h}, \mathcal{C}) = \sum_{m \in \{\text{RNA}, \text{text}, \text{image}\}} \mathbb{I}(m) \cdot \text{Attention}\!\left(\mathbf{h}\mathbf{W}^Q,\, \mathbf{c}_m\mathbf{W}^K_m,\, \mathbf{c}_m\mathbf{W}^V_m\right)
\end{equation}
where $\mathbb{I}(m)$ indicates whether modality $m$ is active and the $\mathbf{W}$ matrices are trainable projections. This design lets the network dedicate separate parameters to each modality, preserving their unique biological characteristics while enabling flexible any-modality conditioning.

\noindent
\textbf{Image and Text Encoding.} We use MUSK\cite{xiang2025vision} to encode image and text conditions into high-quality, clinically relevant representations that guide the generative network.

\noindent
\textbf{Feature Alignment via Teacher Model.}
We align the student SiT's internal features with those of a pretrained pathology encoder (Virchow2\cite{zimmermann2024virchow2}) acting as a teacher. A lightweight CNN projector bridges the two feature spaces:
\begin{equation}
    \mathcal{L}_{\text{align}} = \mathbb{E}_{t, \mathbf{x}_0, \epsilon} \left[ \left\| F_{\text{teacher}}(\mathbf{x}_0) - \text{CNN}_{\text{proj}}\!\left( F_{\text{SiT}}(\mathbf{x}_t, t, \mathcal{C}) \right) \right\|_2^2 \right]
\end{equation}

\noindent
\textbf{Denoising Objective with Auxiliary CLS Loss.}
Beyond patch-level denoising, we add an auxiliary CLS token loss to encourage globally coherent image generation:
\begin{equation}
    \mathcal{L}_{\text{denoise}} = \mathbb{E}_{t, \mathbf{x}_0, \epsilon} \!\left[ \lambda_{\text{patch}} \left\| \mathbf{v}_{\text{patch}} - \mathbf{v}^*_{\text{patch}} \right\|_2^2 + \lambda_{\text{CLS}} \left\| \mathbf{v}_{\text{CLS}} - \mathbf{v}^*_{\text{CLS}} \right\|_2^2 \right]
\end{equation}

\noindent
\textbf{Implementation.} The model is trained with AdamW ($\text{lr}=10^{-4}$, bf16 precision) on 8 H100 GPUs for up to 500K steps. Images are encoded into latents via a frozen VAE ($8\times$ downsampling). Classifier-free guidance uses 10\% condition dropout; an EMA of model weights (decay 0.9999) stabilizes training. Inference uses 250 SDE steps with dynamic guidance annealing from 2.5 to 0.0.

\heading{Application: Text-to-Image and Image-to-Image Generation}

\noindent
\textbf{Text-to-Image.} We evaluate zero-shot generation on 100K held-out captions from PathGen-1.6M\cite{sun2024pathgen} using FID (computed over 25 bootstrap iterations with H-optimus-1\cite{hoptimus1} features) and cosine similarity (image–image via H-optimus-1; text–image via CONCH\cite{lu2024visual}). Baselines include PathLDM\cite{yellapragada2024pathldm}, Stable Diffusion 3.5\cite{esser2024scaling}, and FLUX 2.0\cite{flux-2-2025}.

\noindent
\textbf{Image-to-Image.} Using 100K samples from HISTAI\cite{nechaev2025histai} as references, we report FID and mean cosine similarity (H-optimus-1) against PixCell\cite{yellapragada2025pixcell}, CytoSyn\cite{filiot2025cytosyn}, and SD 3.5\cite{esser2024scaling}.

\heading{Application: Synthetic Data Augmentation}

\noindent\textbf{Few-Shot Classification.} To assess generation utility under data scarcity, we augment 5-shot and 10-shot training sets across five datasets (SkinCancer\cite{skincaner}, PanNuke\cite{gamper2019pannuke}, Unitopatho\cite{barbano2021unitopatho}, LC25000\cite{borkowski2019lung}, SICAPv2\cite{silva2020going}) with \ours-generated images. A ResNet-18\cite{he2016deep} classifier is trained with SGD for 50 epochs; results are averaged over 5 random seeds.

\noindent\textbf{Vision-Language Alignment.} We generate 100K synthetic image-text pairs by prompting \ours with curated pathology descriptions, then fine-tune a CLIP ViT-B/16\cite{radford2021learning} on this corpus using a symmetric InfoNCE loss (AdamW, $\text{lr}=5\times10^{-6}$, 5 epochs). We evaluate Recall@\{10,50\} on image-to-text and text-to-image retrieval on PathMMU\cite{sun2024pathmmu}.

\heading{Application: Spatial Transcriptomics to H\&E Generation}

\noindent
Although \ours is pretrained on bulk RNA-seq, we evaluate transcriptomics-conditioned generation on spatial transcriptomics (ST) data. Both bulk and ST profiles are compressed into the same 331 pathway-level scores, enabling zero-shot transfer of the pretrained priors. Fine-tuning on ST data from MOSAIC refines these priors toward spatially resolved generation. WSIs are tiled into 55$\mu$m spots ($256\times256$ px); paired ST matrices are aligned to extract regional expression profiles per patch. We fine-tune with AdamW ($\text{lr}=10^{-4}$, 4 GPUs, 50K steps). Evaluation uses H-optimus-1 FID and cell-type distribution fidelity via Histoplus\cite{histoplus2024} (14 cell classes).

\heading{Application: Fresh Frozen to FFPE Translation}

\noindent
We address the visual domain gap between Fresh Frozen (FF) and FFPE tissue preparations using an unpaired domain-conditioned pipeline built on \ours, trained on TCGA lung and brain cohorts\cite{ozyoruk2022deep}. Domain identity is injected via text conditioning (``fresh frozen'' or ``ffpe image'') with 10\% dropout for classifier-free guidance.

\noindent
\textbf{Structure-Guided Inference.} To preserve spatial architecture during translation, we combine DDIM inversion with cross-domain attention injection. The FF image is inverted to $z_T$ under the source prompt $c_{\text{FF}}$, then decoded under $c_{\text{FFPE}}$:
\begin{align}
z_T &= \text{DDIM-Invert}(z_0,\, c_{\text{FF}},\, z_{\text{cls}})\\
\hat{z}_0 &= \text{ODE-Solve}(z_T,\, c_{\text{FFPE}},\, z_{\text{cls}})
\end{align}
To maintain structural fidelity, self-attention maps $A_t^\ell$ from a parallel source-conditioned reconstruction pass are injected into the target pass, constraining the generated image to inherit the cellular layout of the input. We benchmark against CycleGAN\cite{zhu2017unpaired}, CUT\cite{park2020contrastive}, and AI-FFPE\cite{ozyoruk2022deep} using FID and clinical expert assessment.

\heading{Application: H\&E-to-IHC Translation}

\noindent
We evaluate on HER2Match\cite{HER2match_data} (co-registered H\&E/HER2 pairs) and IHC4BC\cite{Akbarnejad2023PredictingKE} (serial-section H\&E with Ki67, PR, ER, HER2 markers). Before fine-tuning \ours for pixel-level IHC synthesis, we bridge the modality gap in MUSK latent space via flow matching. A 6-layer MLP $v_\theta$ learns a continuous flow from H\&E embeddings $z_{\text{HE}}$ to IHC embeddings $z_{\text{IHC}}$:
\begin{equation}
    \mathcal{L}_{\text{FM}} = \mathbb{E}_{t,\, z_0,\, z_1} \left[ \left\| v_\theta(z_t, t) - (z_1 - z_0) \right\|_2^2 \right]
\end{equation}
At inference, the translated embedding $\hat{z}_{\text{IHC}}$ conditions \ours to generate the target IHC image. Clinical utility is assessed via gray-zone AUC\cite{Akbarnejad2023PredictingKE} for ER, PR, Ki67, and HER2. We compare against CycleGAN\cite{zhu2017unpaired} and CUT\cite{park2020contrastive} using FID and KID.

\heading{Application: H\&E-to-mIF Translation}

\noindent
We fine-tune \ours on the ORION colorectal cancer dataset\cite{lin2023high} (41 WSIs, 16 protein markers). H\&E conditioning is applied through two complementary mechanisms: (1)~\textbf{spatial concatenation} of the H\&E VAE latent $z_{\text{struct}}$ with the noisy mIF latent at each timestep, providing a structural template; and (2)~\textbf{semantic cross-attention} using MUSK patch embeddings $c_{\text{sem}}$ to modulate generation based on tissue pathology. Multiplex channels are grouped into 3-channel subsets and encoded with the VAE independently. The model is trained with flow-matching velocity prediction loss using BF16 mixed precision with HED-based H\&E color augmentation. We evaluate Patch PCC and Slide-level PCC against GigaTIME\cite{valanarasu2025multimodal} and HistoPlexer\cite{andani2025histopathology}.

%% file: appendix/figs.tex

\begin{figure*}[!t]
\centering
\includegraphics[width=0.9\linewidth]{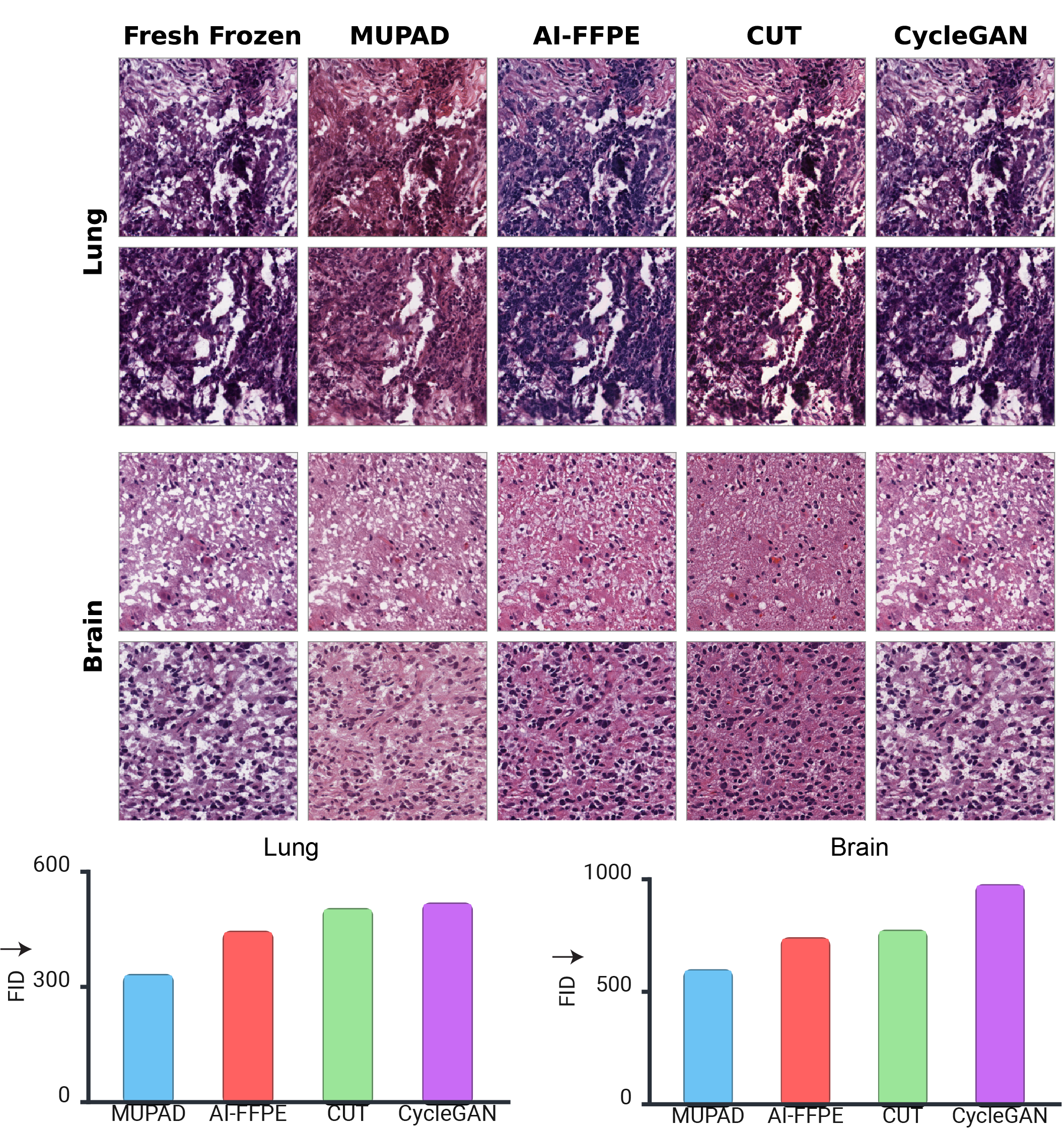}
\caption{
    \textbf{Fresh frozen to FFPE image translation results.}    Visual comparison of translation quality on Lung and Brain tissues. \ours\ successfully suppresses cryo-artifacts typical of fresh frozen sections while synthesizing realistic H\&E staining characteristics. Baselines often suffer from saturation (AI-FFPE) or residual noise (CUT, CycleGAN).    Quantitative evaluation showing \ours\ achieves the lowest FID (lower is better).
}
\label{fig:fig4-ff2ffpe}
\end{figure*}

\begin{figure*}[!h]
    \centering
    \includegraphics[width=0.98\linewidth]{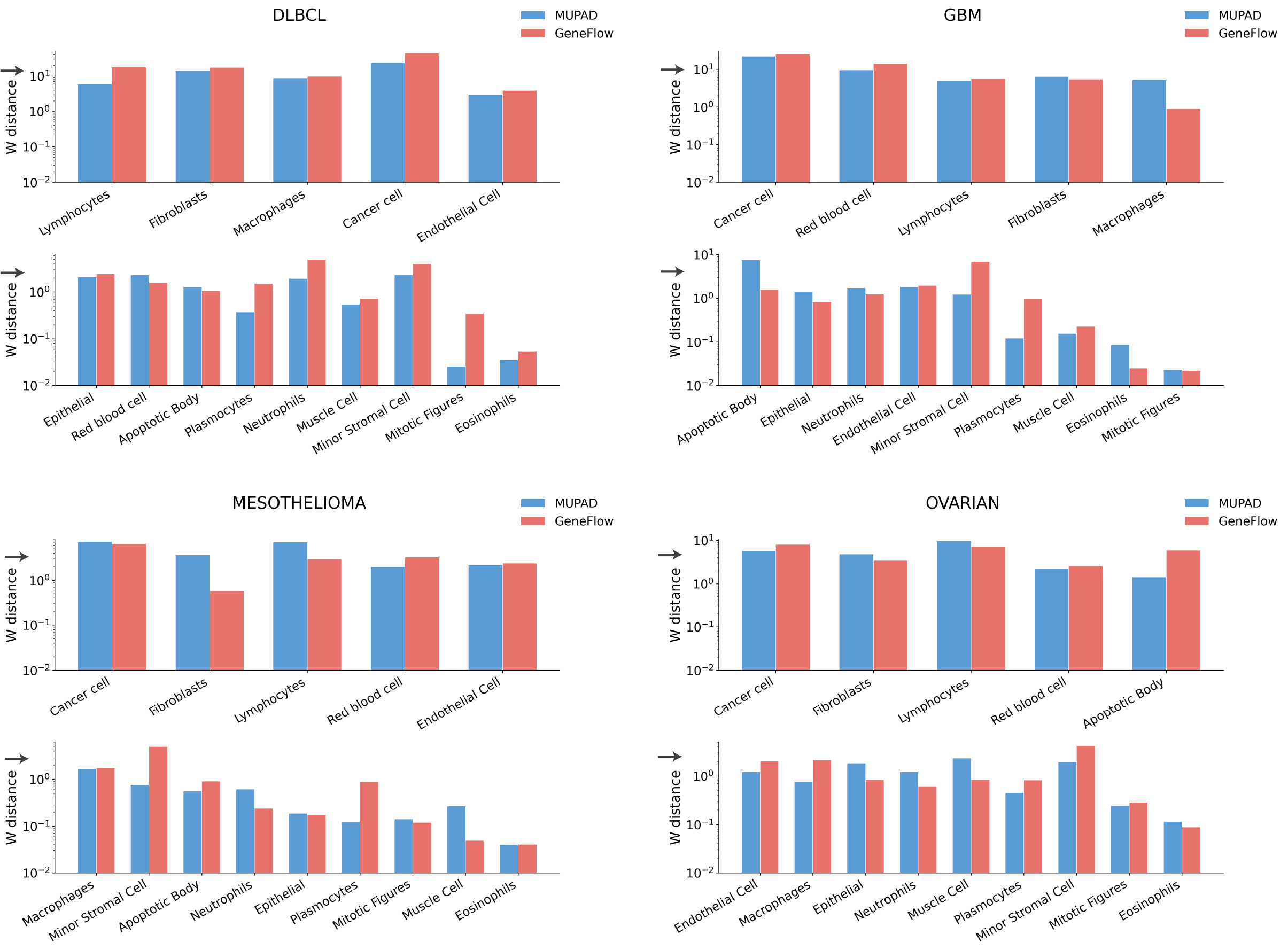}
    \caption{ Generating H\&E images from spatial transcriptomic. Synthetic samples preserve spatially resolved cell-type distributions for DLBCL, GBM, mesothelioma, and ovarian. The Wasserstein distance (W) quantifies the distribution alignment of real and synthetic. Lower is better. 
    }
    \label{fig:extended_st2image}
\end{figure*}



%% file: appendix/tables.tex
\begin{table}[t]
\caption{\textbf{Comparison of different methods and supported tasks.} FID values are reported relative to MUPAD (1.00), where lower is better.}
\centering
\small 
\begin{tabular}{lccccccc}
\toprule
& Image-to-image & Text-to-image & RNA-to-image & FF-to-FFPE & \multicolumn{2}{c}{HE-to-IHC} & HE-to-mIF \\
\cmidrule(lr){6-7}
& (Rel. FID$\downarrow$) & (Rel. FID$\downarrow$) & (Rel. FID$\downarrow$) & (Rel. FID$\downarrow$) & Rel. FID$\downarrow$ & AUC$\uparrow$ & (Pearson$\uparrow$) \\
\midrule
\rowcolor{LightGray} \textbf{MUPAD} & 1.00 & 1.00 & 1.00 & 1.00 & 1.00 & 0.977 & 0.464 \\
PixCell     & 1.97  & \xmark & \xmark & \xmark & \xmark & \xmark & \xmark \\
SDv3.5      & 4.43  & 2.80  & \xmark & \xmark & \xmark & \xmark & \xmark \\
FLUX.2      & 2.46  & 2.45  & \xmark & \xmark & \xmark & \xmark & \xmark \\
PathLDM     & \xmark & 1.79  & \xmark & \xmark & \xmark & \xmark & \xmark \\
GeneFlow    & \xmark & \xmark & 1.30  & \xmark & \xmark & \xmark & \xmark \\
AI-FFPE     & \xmark & \xmark & \xmark & 1.28  & \xmark & \xmark & \xmark \\
CUT         & \xmark & \xmark & \xmark & 1.38  & 2.16  & 0.902 & \xmark \\
CycleGAN    & \xmark & \xmark & \xmark & 1.62  & 1.70  & 0.873 & \xmark \\
GigaTIME    & \xmark & \xmark & \xmark & \xmark & \xmark & \xmark & 0.339 \\
HistoPlexer & \xmark & \xmark & \xmark & \xmark & \xmark & \xmark & 0.192 \\
\bottomrule
\end{tabular}
\label{tab:tasks-comparison}
\end{table}